%% file: ms.tex
\documentclass[11pt,letterpaper]{article}
\usepackage{emnlp2017}
\usepackage{times}
\usepackage{amsmath}
\usepackage{latexsym}
\usepackage{enumitem}
\usepackage{graphicx}
\usepackage{multicol}
\usepackage{lipsum}
\usepackage{subcaption}
\usepackage{multirow}
\graphicspath{{images/}}
\usepackage{algorithm}
\usepackage[noend]{algpseudocode}

\algdef{SE}[DOWHILE]{Do}{doWhile}{\algorithmicdo}[1]{\algorithmicwhile\ #1}%

\newlist{inlinelist}{enumerate*}{1}
\setlist*[inlinelist,1]{%
  label=(\roman*),
}

\emnlpfinalcopy

\def\emnlppaperid{022}

\title{Unsupervised Aspect Term Extraction with B-LSTM \& CRF using Automatically Labelled Datasets}

\author{Athanasios Giannakopoulos,~~Claudiu Musat,~~Andreea Hossmann
        \\ { \and \bf Michael Baeriswyl}
        \\ Artificial Intelligence and Machine Learning Group | Swisscom AG 
        \\ {\tt firstName.lastName@swisscom.com}
         }

\begin{document}

\maketitle

\input{sections/abstract}

\section{Introduction} \label{sec:intro}
\input{sections/introduction}

\section{Related Work} \label{sec:related work}
\input{sections/related_work}

\section{Supervised Aspect Term Extraction} \label{sec:supervised ate}
\input{sections/supervised_ate}

\section{Unsupervised Aspect Term Extraction} \label{sec:unsupervised ate}
\input{sections/unsupervised_ate}

\section{Experiments and Results} \label{sec:experiments}
\input{sections/experiments}

\section{Conclusion and Future Work} \label{sec:conclusion}
\input{sections/conclusion}

% \section{Future Work} \label{sec:future work}
% \input{sections/futute_work}

%\section*{Acknowledgments}
%\input{sections/acknowledgements}

\bibliography{emnlp2017}
\bibliographystyle{emnlp_natbib}

\end{document}

%% file: sections/abstract.tex
\begin{abstract}
Aspect Term Extraction (ATE) identifies opinionated aspect terms in texts and is one of the tasks in the SemEval Aspect Based Sentiment Analysis (ABSA) contest. The small amount of available datasets for supervised ATE and the costly human annotation for aspect term labelling give rise to the need for unsupervised ATE. In this paper, we introduce an architecture that achieves top-ranking performance for supervised ATE. Moreover, it can be used efficiently as feature extractor and classifier for unsupervised ATE. Our second contribution is a method to automatically construct datasets for ATE. We train a classifier on our automatically labelled datasets and evaluate it on the human annotated SemEval ABSA test sets. Compared to a strong rule-based baseline, we obtain a dramatically higher \textit{F-score} and attain precision values above 80\%. Our unsupervised method beats the supervised ABSA baseline from SemEval, while preserving high precision scores.
\end{abstract}

%% file: sections/introduction.tex
For many years now, companies are offering users the possibility of adding reviews in the form of sentences or small paragraphs. Reviews can be beneficial for both customers and companies. On the one hand, people can make better decisions by getting insights about available products and solutions. One the other hand, companies are interested in understanding how and what customers think about their products, which helps in employing marketing solutions and correction strategies. To this end, performing an automated analysis of the user opinions becomes a crucial issue.

\noindent
Performing sentiment analysis to detect the overall polarity of a sentence or paragraph comes with two major disadvantages. First, sentiment analysis on sentence (or paragraph) level does not fulfill the purpose of getting more accurate and precise information. The polarity refers to a broader context, instead of pinpointing specific targets. Secondly, many sentences or paragraphs contain opposing polarities towards distinct targets, making it impossible to assign an accurate overall polarity.

\noindent
The need for identifying aspect terms and their respective polarity gave rise to the Aspect Based Sentiment Analysis (ABSA), where the task is first to extract aspects or features of an entity (i.e. Aspect Term Extraction or ATE\footnote{Also known as Opinion Term Extraction (OTE).}) from a given text, and second to determine the sentiment polarity (SP), if any, towards each aspect of that entity. The importance of ABSA led to the creation of the ABSA task in the SemEval\footnote{The SemEval (Semantic Evaluation) contest is an ongoing series of evaluations of computational semantic analysis systems.} contest in 2014~\cite{semeval2014}.

\noindent
Supervised ATE using human annotated datasets leads to high performance for aspect term detection on unseen data, however it has two major drawbacks. First, the size of the labelled datasets is quite small, reducing the performance of the classifiers. Second, human annotation is a very slow and costly procedure.
% %\begin{enumerate}[topsep=1pt, itemsep=-1ex, partopsep=1ex, parsep=1ex]
% \begin{inlinelist}
%     \item the size of the labelled datasets is quite small, reducing the performance of the classifiers and
%     \item human annotation is a very slow and costly procedure.
% \end{inlinelist}
% %\end{enumerate}

\noindent
The drawbacks of supervised ATE can be tackled using unsupervised ATE. The size of the datasets can be significantly increased using targets from publicly available reviews (e.g. \textit{Amazon} or \textit{Yelp}). Reviews are opinion texts and contain plenty of opinionated aspect terms, which makes them perfect candidates for constructing new datasets for ATE. With respect to the second drawback, an automated data labelling process with high precision can replace the slow and error-prone human annotation procedure.

\noindent
We innovate by performing ATE starting from opinion texts (e.g. reviews). This is a completely unsupervised task since there are no labels to explicitly pinpoint that certain tokens of the text are aspect terms. Reviews may contain labels (e.g. number of stars in a 1-5 star rating system) which are related to their overall polarity. However, such labels are not useful for ATE.

\noindent
In this work, we present a classifier, which can be used for feature extraction and aspect term detection for both unsupervised and supervised ATE. We validate its suitability for ATE by achieving top-ranking results for supervised ATE using the SemEval-2014 ABSA task datasets\footnote{The SemEval ABSA datasets contain human annotation for ATE for both the laptop and the restaurant domains only in 2014.}. Then, we use it for unsupervised ATE.

\noindent
Moreover, we contribute by introducing a new, completely automated, unsupervised and domain independent method for annotating raw opinion texts. Then, we use our classifier to perform unsupervised ATE by training it on the automatically labelled datasets obtained with our method. Against all expectations, our unsupervised method beats the supervised ABSA baseline from SemEval-2014, while achieving high precision scores. The latter is very important for unsupervised techniques since we wish to extract non-noisy aspect terms, i.e. minimize the number of false positives.

\noindent
The rest of this paper is organized as follows. Section~\ref{sec:related work} presents the related work for ATE. Our approach for supervised and unsupervised ATE is described in Sections~\ref{sec:supervised ate} and~\ref{sec:unsupervised ate} respectively. Section~\ref{sec:experiments} presents our experiments and results for both supervised and unsupervised ATE. Finally, Section~\ref{sec:conclusion} focuses on our conclusions and future work.

%% file: sections/related_work.tex
% Research in the area of both supervised and unsupervised ATE has flourished after the creation of the SemEval ABSA task in 2014. The winners of the SemEval-2014 ABSA contest~\cite{winners2014} used supervised methods for ATE. They utilized two categories of features: 
% \begin{inlinelist}
%     \item general features and
%     \item features generated from open/external sources.
% \end{inlinelist}
% Features of the first category are created using the provided training set and are very similar to those used in traditional Name Entity Recognition (NER) systems~\cite{nerFeatures}. Features of the second category can emerge by using external sources, such as the WordNet lexicographer files~\cite{wordnet} and word clusters derived using Brown clustering~\cite{brown} or K-means\footnote{\url{https://en.wikipedia.org/wiki/K-means_clustering}}. Other types of features used for ATE are presented in~\cite{nlang2015} and include the use of gazetteers~\cite{gazetteers} and word embeddings~\cite{we}.~\citet{nlang2016} use the probability output of an Recurrent Neural Network (RNN) to further enrich the feature space.

Research in the area of both supervised and unsupervised ATE has flourished after the creation of the SemEval ABSA task in 2014. The winners of the SemEval-2014 ABSA contest~\cite{winners2014} use supervised methods for ATE. They extract features, similar to those used in traditional Name Entity Recognition (NER) systems~\cite{nerFeatures} using the provided training sets. Moreover, they exploit external sources, such as the WordNet lexicographer files~\cite{wordnet} and word clusters (e.g. Brown clusters~\cite{brown} or K-means\footnote{\url{https://en.wikipedia.org/wiki/K-means_clustering}}).~\citet{nlang2015} suggest using gazetteers~\cite{gazetteers} and word embeddings~\cite{we} for ATE.~\citet{nlang2016} use the probability output of an Recurrent Neural Network (RNN) to further enrich the feature space.

\noindent
Independently of the feature extraction techniques, supervised ATE is treated as a sequential labelling task. Top-ranking participants in the SemEval ABSA contest use Conditional Random Fields (CRF) or Support Vector Machine (SVM) as sequential labelling classifiers~\cite{winners2014, nlang2015, ihs, xrce}.

\noindent
There is also related work with respect to unsupervised ATE. ~\citet{liuRules} exploit syntactic rules to automatically detect aspect terms.~\cite{v3, spanishPaper} use a graph representation to describe the interactions between aspect terms and opinion words in raw text. Graph nodes are ranked using PageRank and high-ranked nodes are used to create a set of aspect terms. Then, they use this set to annotate unseen data by simply performing exact or lemma matching.

\noindent
Systems similar to~\cite{absaCzech, pathEmbeddings, Poria201642} perform semi-supervised ATE since they use human annotated datasets for training but enrich their feature space using features extracted by exploiting large unlabelled corpora.~\citet{greek_paper} present a method for constructing new datasets for ATE, however they use non-standard evaluation metrics. Finally, systems like~\cite{W2VLDA} focus on classifying the aspect terms into categories. We do not compare against such systems, since they do not perform the same task and are not equivalent to ours.

\noindent
In all but one\footnote{\citet{greek_paper} use a non-standard definition of precision and recall.} aforementioned cases, the evaluation of the model is performed using the \textit{F-score}, as defined in~\cite{semeval2014}. In case of unsupervised ATE, achieving higher precision is more important than higher recall as highlighted in~\cite{liuRules}. 

\noindent
We perform both supervised and unsupervised ATE using a model that utilizes continuous word representations and performs feature extraction and sequential labelling simultaneously while training. In case of supervised ATE, the training datasets are those of the SemEval ABSA task (human annotated). In case of unsupervised ATE, we annotate raw opinion texts (e.g. reviews) with a completely automated and unsupervised process, which we introduce. To the best of our knowledge, we are the first to train a classifier using an automatically labelled dataset and perform evaluation on human annotated datasets.

%% file: sections/supervised_ate.tex
The ATE task can be modelled as a token-based classification task, where labels are assigned to the tokens of a sequence, depending on whether they are aspect terms or not. For supervised ATE, we apply a classification pipeline that consists of 3 steps:
\begin{inlinelist}
    \item data preprocessing,
    \item model training and
    \item model evaluation.
\end{inlinelist}

\noindent
The feature extraction is performed from a two-layer bidirectional long short-term memory (B-LSTM) network while the model is training, similar to the way a Convolutional Neural Network (CNN) extracts features while performing image classification. Therefore, we do not explicitly include this step in the aforementioned pipeline. 

\subsection{Data Preprocessing}
We break down each sentence into tokens using the spaCy parser\footnote{\url{https://spacy.io/docs/}} and follow the traditional IOB format (short for Inside, Outside, Beginning) for sequential labelling. Tokens that represent the aspect terms of the sentence are labelled with B. In case an aspect term consists of multiple tokens, the first token receives the B label and the rest receive the I label. Tokens that are not aspect terms are labelled with O. Given the sentence "The internal speakers are amazing." with target "internal speakers", the labelling would be as follows: (The$|$O) ~ (internal$|$B) ~ (speakers$|$I) ~ (are$|$O) ~ (amazing$|$O) ~ (.$|$O). 

\subsection{Classifier Architecture}
\begin{figure*}[t]
    \centering
    \includegraphics[width=0.7\textwidth, height=5.5cm]{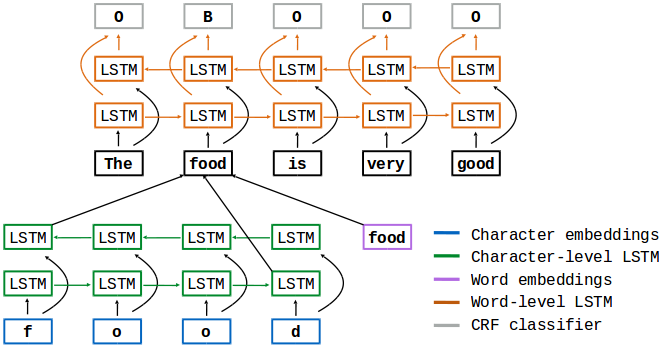}
    \caption{Sequential labelling using B-LSTM \& CRF classifier.}
    \label{fig:neuroate}
\end{figure*}

We employ a two-layer B-LSTM to extract features for each token, which are used by a CRF for token-based classification. Features are created by exploiting the word morphology and the structure of the sentence. The architecture is depicted in Fig.~\ref{fig:neuroate} and is inspired by the NER system presented in~\cite{neuroner}. However, we employ LSTM cells and use word embeddings from fastText\footnote{https://github.com/facebookresearch/fastText}.

\noindent
\textbf{First B-LSTM layer:} Randomly initialized character embeddings of each token are given as input into the first B-LSTM layer, which aims at learning new word embeddings. The first and second directions (left $\rightarrow$ right and left $\leftarrow$ right) of the first B-LSTM layer are responsible for learning word embeddings by exploiting the prefix and the suffix of each token respectively.

\noindent
\textbf{Second B-LSTM layer:} For each token of a sentence, we create a feature vector by combining  
\begin{inlinelist}
    \item the extracted word embeddings from the first B-LSTM layer and
    \item pre-trained word embeddings using fastText.
\end{inlinelist}
These feature vectors are given as input to the second B-LSTM layer, which extracts a feature vector for each token by exploiting the structure of the sentence. Similar to the first B-LSTM layer, the first and second directions are responsible for extracting features using the previous and the next tokens of each word.

\noindent
\textbf{CRF layer:} The final layer uses the extracted feature vectors in order to perform sequential labelling.

%% file: sections/unsupervised_ate.tex
The human annotation process | required to identify aspect terms in small sentences and construct datasets for supervised ATE | comes at a high cost, mainly for two reasons:
\begin{enumerate}[topsep=1pt, itemsep=-1ex, partopsep=1ex, parsep=1ex]
    \item Human annotated datasets typically consist of a few thousand sentences\footnote{The datasets of the SemEval ABSA task consist of approximately 3000 sentences for English.} extracted from large corpora of domain-specific reviews. The \textbf{small amount of data} reduces the performance of classifiers.
    \item \textbf{Human annotation} is very slow, costly and risky. Annotators may introduce noise in the datasets by labelling words incorrectly, either because they are sloppy workers or because they do not know exactly what aspect terms are. For example, given the sentence "Works well, and I am extremely happy to be back to an apple OS.", human annotators may consider the word "works" as a target\footnote{Example taken from the golden annotated dataset for laptop reviews of the SemEval-2014 ABSA task.}. However, aspect terms are nouns and noun phrases~\cite{liuRules}, therefore the verb "works" should not be considered as a target. 
\end{enumerate}

\noindent
We employ unsupervised ATE in order to overcome both problems. We tackle the first problem by using large datasets of opinion texts (e.g. reviews). Such datasets are ideal for ATE since they contain a plethora of opinionated aspect terms. In order to tackle the second problem, we introduce and use an automated and unsupervised method for labelling the tokens of the aforementioned datasets using the IOB format. We consider only noun and noun phrases as candidate aspect terms and focus on token labelling with high precision in order to reduce falsely annotated aspect terms. In that way, we minimize the cost, the time and the mistakes introduced by the human annotation process.

\noindent
We use the publicly available datasets of \textit{Amazon} and \textit{Yelp} for laptop and restaurant reviews respectively and perform some data cleaning such as removing URLs from the reviews. 
% Our pipeline for unsupervised ATE consists of
% \begin{inlinelist}
%     \item automated data labelling,
%     \item model training and
%     \item model evaluation on human annotated datasets.
% \end{inlinelist}

% \subsection{Automated Data Labelling in a Nutshell}
% Given a large domain-specific review corpus, Algorithm~\ref{alg:unsupervisedATE} can be used to automatically label the raw opinion text. Its steps are as follows:
% \begin{enumerate}[topsep=1pt, itemsep=-1ex, partopsep=1ex, parsep=1ex]
%     \item We run AutoPhrase on a given large corpus (preferably domain dependent) in order to extract the quality phrases. The list of quality phrases is pruned according to a desired threshold value.
%     \item We iterate through all the sentences in the corpus. Tokens of sentences that obey certain syntactic rules are labelled as aspect terms.
%     \item We repeat the aforementioned procedure for potential multi-word aspect terms.
%     \item We label each sentence following the IOB labelling format.
% \end{enumerate}

\subsection{Automated Data Labelling} \label{ssec:data labelling}
Using raw opinion texts (e.g. reviews) for ATE is a completely unsupervised task since there are no labels to explicitly pinpoint that certain tokens of the text are aspect terms. Reviews frequently contain labels (e.g. number of stars in a 1-5 star rating system) related to their overall polarity but these are not useful for ATE.

\noindent
Our goal is to label each token of the unlabelled opinion texts in an automated way using the IOB format with unsupervised methods. While labelling aspect terms, we focus on high precision, a property that guarantees that the resulting datasets will contain as little noisy aspect terms as possible. The importance of high precision is also highlighted in~\cite{liuRules}, where authors construct syntactic rules primarily by focusing on this criterion.

\noindent
Algorithm~\ref{alg:unsupervisedATE} describes the automated data labelling method. First, we create a list of quality phrases and prune it using a desired threshold value. Then, we iterate through all sentences and annotate tokens that obey certain syntactic rules as aspect terms. We repeat this procedure for multi-word aspect terms and finally label the tokens using the IOB format.

\begin{algorithm}
\caption{Automated Data Labelling} \label{alg:unsupervisedATE}
\begin{algorithmic}[1]
\State \emph{qual\_phrases = run\_autophrase(corpus)}
\State \emph{candidates = prune(qual\_phrases, $q_{th}$)}
\For{sentence \textbf{in} corpus}
    \State labels = []
        \For{token \textbf{in} sentence}
            \If {token \textbf{in} candidates}
                \State \emph{l = get\_label(token, rules, lexicon)}
                \State \emph{labels.append(l)}
            \EndIf
        \EndFor
    \State \emph{assign\_iob\_tags(sentence, labels)}
\EndFor
\end{algorithmic}
\end{algorithm}

\subsubsection{Quality Phrase List}
We start by building a sorted list of the form $(\textit{quality phrase}, q)$, where $q \in [0,1]$ represents the quality value of each phrase. The quality phrases~|~which we use as candidate aspect terms~|~are \textit{n-grams} that appear in the raw review corpora and exceed a minimum support threshold\footnote{Support is an indication of how frequently the \textit{n-gram} appears in the dataset.}. The list of quality phrases is built by applying the AutoPhrase algorithm~\cite{autophrase} on the review datasets for laptops and restaurants. The quality of each phrase is determined via a classification task with decision trees that takes into account a list of high quality phrases using \textit{Wikipedia}. The values of the features (e.g. \textit{tf-idf}) used in the decision trees to predict the quality of each phrase are more informative when the provided corpora are domain dependent. Therefore, we apply AutoPhrase on each dataset separately, rather than combining the two datasets.

\noindent
The extracted quality phrases, together with a set of syntactic rules, contribute in the automated data labelling process, which is based on 3 pillars:
\begin{enumerate}[topsep=1pt, itemsep=-1ex, partopsep=1ex, parsep=1ex]
    \item a sentiment lexicon
    \item a pruned list of quality phrases
    \item syntactic rules able to capture aspect terms
\end{enumerate}
Existing ATE systems~\cite{v3}, although unsupervised, exploit also syntactic rules derived from supervised tools (e.g. parsers). Moreover, they require domain-dependent human input (e.g. seed words) to perform double-propagation. We avoid that by using a sentiment lexicon.

\subsubsection{Sentiment Lexicon} 
In many cases, aspect terms have modifiers (e.g. "This is a great screen") or are objects of verbs (e.g. "I love the screen of this laptop") that express a sentiment. Therefore, we make use of a sentiment lexicon\footnote{We use the sentiment lexicon of Bing Liu: \\ \url{https://www.cs.uic.edu/~liub/FBS/sentiment-analysis.html}}, which is necessary in order to perform a look-up on whether modifiers and verbs express a sentiment or not.

\subsubsection{Pruned Quality Phrases} \label{sssec: pruned quality phrases}
We prune our quality phrases since they contain both true and noisy aspect term candidates. More concretely, we filter the list of quality phrases in order to keep \textit{n-grams} with a quality above a certain threshold. 

\noindent
We present an example to show the value of the pruning step. The list of quality phrases extracted using the \textit{Amazon} review dataset on laptops contains the \textit{1-gram} "couch" and the \textit{2-gram} "touch pad" with quality 0.67 and 0.95 respectively. However, the presence of the word "couch" as an aspect term in laptop reviews is completely arbitrary. Therefore, we prune the list of quality phrases using an empirical quality threshold of $q_{th}=0.7$ and $q_{th}=0.6$ for the laptop and restaurant domain respectively. We set these thresholds manually after inspecting the lists of quality phrases and detecting the quality value under which a lot of domain-irrelevant candidate aspect terms appear. 

\noindent
While the pruning step removes irrelevant phrases, as shown above, it also means that \textit{n-grams} such as "set up", which are true aspect term candidates are removed from the list due to low quality ($q_{set \ up} = 0.32$). However, reducing noisy aspect term candidates (e.g. "couch" with $q=0.67$) is more important than keeping all aspect term candidates since we wish to annotate aspect terms with high precision.

\noindent
We can make the data labelling method completely automated by setting a fixed quality threshold $q_{th}$ for pruning the list of quality phrases. We highlight that a fixed threshold of $q_{th}=0.7$ leads to a good~|~but not optimal~|~trade-off between high precision values and good \textit{F-score} for ATE.

\begin{table*}[]
\centering
\resizebox{\textwidth}{!}{
\begin{tabular}{l|c|c}
\multicolumn{1}{c|}{\textbf{Rules}} & \textbf{Example} & \multicolumn{1}{c}{\textbf{Extracted Targets}} \\ \hline \hline
\begin{tabular}[c]{@{}l@{}} $depends(dobj, t_i, t_j)$ \textbf{and} $opinion\_word(t_j)$ \\ \textbf{then} $mark\_target(t_i)$ \end{tabular} & I like the screen & \multicolumn{1}{c}{screen} \\ \hline
\begin{tabular}[c]{@{}l@{}} $depends(nsubj, t_i, t_j)$ \textbf{and} $depends(acomp, t_k, t_j)$ \\  \textbf{and} $opinion\_word(t_k)$ \textbf{then} $mark\_target(t_i)$ \end{tabular} & The internal speakers are amazing & internal speakers \\ \hline
\begin{tabular}[c]{@{}l@{}} $depends(nsubj, t_i, t_j)$ \textbf{and} $depends(advmod, t_j, t_j)$ \\ \textbf{and} $opinion\_word(t_k)$ \textbf{then} $mark\_target(t_i)$ \end{tabular} & The touchpad works perfectly & touchpad \\ \hline
\begin{tabular}[c]{@{}l@{}} $depends(pobj~\mathbf{or}~dobj, t_i, t_j)$ \textbf{and} $depends(amod, t_k, t_i)$ \\ \textbf{and} $opinion\_word(t_k)$ \textbf{then} $mark\_target(t_i)$ \end{tabular} & This laptop has great price & price \\ \hline
\begin{tabular}[c]{@{}l@{}} $depends(cc~\mathbf{or}~conj, t_i, t_j)$ \textbf{and} $is\_aspect(t_j)$ \\ \textbf{then} $mark\_target(t_i)$ \end{tabular} & Screen and speakers are awful & \begin{tabular}[c]{@{}c@{}}screen \\ speakers\end{tabular} \\ \hline
\begin{tabular}[c]{@{}l@{}} $depends(compound, t_i, t_j)$ \textbf{and} $is\_aspect(t_j)$ \\ \textbf{then} $mark\_target(t_i)$ \end{tabular} & The wifi card is not good & wifi card 
\end{tabular}
}
\caption{Syntactic rules for aspect term extraction.}
\label{tab:syntactic rules}
\end{table*}

\subsubsection{Syntactic Rules for ATE} The pruned quality phrases and the sentiment lexicon are combined with syntactic rules in order to extract aspect terms from sentences. Before applying any syntactic rule, we validate if a token is a potential aspect term by checking if it
\begin{inlinelist}
    \item is not a stopword,
    \item is present in the pruned quality phrases and
    \item has a POS tag that is present in [NOUN, PRON, PROPN, ADJ, ADP, CONJ].
\end{inlinelist}
Table~\ref{tab:syntactic rules} tabulates all rules used for ATE and gives examples of reviews with the respective extracted aspect terms. For simplicity, we adopt a syntactic rule notation similar to the one used in~\cite{liuRules}. The functions used in Table~\ref{tab:syntactic rules} have the following interpretation:
\begin{itemize}[topsep=1pt, itemsep=-1ex, partopsep=1ex, parsep=1ex]
    \item $depends(d,t_i,t_j)$ is true if the syntactic dependency between the tokens $t_i$ and $t_j$ is $d$.
    \item $opinion\_word(t_i)$ is true if the token $t_i$ is in the sentiment lexicon.
    \item $mark\_target(t_i)$ means that we mark the token $t_i$ as aspect term.
    \item $is\_aspect(t_i)$ is true if the token $t_i$ is already marked as aspect term.
\end{itemize}

\subsubsection{Language and Domain Adaptation}

The automated data labelling method requires adaptation in order to be used in different languages. More concretely, we need to adapt 
\begin{inlinelist}
    \item the syntactic rules of Table~\ref{tab:syntactic rules}, 
    \item the sentiment lexicon and
    \item the tools required from Autophrase (e.g. part-of-speech tagger)
\end{inlinelist}
to the target language.

\noindent
We can use the automated data labelling method for ATE dataset construction in a completely domain-independent fashion. To do so, we only need to set the pruning threshold $q_{th}$ of the quality phrase list to a fixed value (Section \ref{sssec: pruned quality phrases}). Our experiments reveal that setting $q_{th}=0.7$ results in a good trade-off between high precision and \textit{F-score}, independently of the laptop or the restaurant domain.

\subsection{Model Training and Evaluation} \label{unsupervised training}
We train a B-LSTM \& CRF classifier to perform unsupervised ATE for both domains using the automatically labelled datasets constructed in Section~\ref{ssec:data labelling}. The classifier is evaluated on the human annotated test datasets of the SemEval-2014 ABSA contest.

%% file: sections/experiments.tex
% We perform experiments for supervised and unsupervised ATE in the laptop and the restaurant domain and evaluate our classifier using the CoNLL\footnote{\url{http://www.cnts.ua.ac.be/conll2003/}} \textit{F-score}, defined in Eq.~\ref{F-score}.
% \noindent
% \begin{equation} \label{F-score}
%     F_1 = \frac{2 \cdot P \cdot R}{P+R}
% \end{equation}
% $P$ and $R$ stand for precision and recall and are given by Eq.~\ref{precision} and~\ref{recall} respectively.
% \vspace{-8mm}

% \begin{multicols}{2}
%     \noindent
%     \begin{equation} \label{precision}
%         P = \frac{|S \cap G|}{|S|}
%     \end{equation}
%     \begin{equation} \label{recall}
%         R = \frac{|S \cap G|}{|G|}
%     \end{equation} 
% \end{multicols}

% \noindent
% Here, $S$ is the set of aspect terms that our system returns for all the test sentences and $G$ is the set of the correct aspect terms~\cite{semeval2014}.

% \noindent
% Compared to other supervised learning methods, we reach the performance of SemEval-2014 ABSA winners in the restaurant domain. For laptops, our supervised system exceeds the best \textit{F-score} of SemEval-2014 ABSA contest by approximately 3\%. With respect to unsupervised ATE, our technique achieves
% \begin{inlinelist}
%     \item very high precision and
%     \item an \textit{F-score} that exceeds the supervised baseline of the SemEval ABSA.
% \end{inlinelist}

We perform experiments for supervised and unsupervised ATE in the laptop and the restaurant domain and evaluate our classifier using the CoNLL\footnote{\url{http://www.cnts.ua.ac.be/conll2003/}} \textit{F-score}. Compared to other supervised learning methods, we reach the performance of SemEval-2014 ABSA winners in the restaurant domain. For laptops, our supervised system exceeds the best \textit{F-score} of the SemEval-2014 ABSA contest by approximately 3\%. With respect to unsupervised ATE, our technique achieves
\begin{inlinelist}
    \item very high precision and
    \item an \textit{F-score} that exceeds the supervised baseline of the SemEval ABSA.
\end{inlinelist}

\subsection{Experiments for Supervised ATE} \label{ssec:supervised ate}
For supervised learning, we perform experiments using the human annotated training and test sets provided by the SemEval-2014 ABSA contest for the laptop and restaurant domain. Our classifier uses the B-LSTM \& CRF architecture presented in Fig.~\ref{fig:neuroate} and its implementation is based on~\cite{neuroner_implementation}. 

\begin{figure}[]
    \centering
    \includegraphics[width=0.47\textwidth]{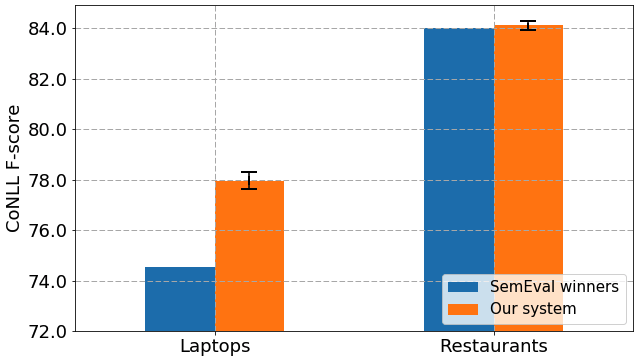}
    \caption{Results for supervised ATE using the B-LSTM \& CRF architecture. We compare against the winners of the SemEval-2014 ABSA contest.}
    \label{fig:supervised experiments}
\end{figure}

\noindent
We use a random 80-20\% split on the original training set of SemEval-2014 ABSA contest in order to create a new train and validation set. We keep the test set for our final evaluation. For most of the parameters, we use the default values of~\cite{neuroner_implementation}. However, we use the \textit{adam} optimizer with learning rate $\alpha=0.01$ and a batch size of 64. Moreover, we use the pre-trained word embeddings of fastText.

\noindent
We train the classifier using the reduced training set for a maximum number of 100 epochs. After each epoch, we evaluate our model using the CoNLL \textit{F-score} on the validation set. Moreover, we use early stopping with a patience of 20 epochs. This means that the experiment terminates earlier if the CoNLL \textit{F-score} of the validation set does not improve for 20 consecutive epochs. At the end of each experiment we choose the model of the epoch that gives the best performance on the validation set and make predictions on the test set. We repeat the aforementioned procedure for 50 experiments and present the experimental results for both domains in Fig~\ref{fig:supervised experiments}. 

\noindent
The \textit{F-score} of the SemEval-2014 ABSA winners is 74.55 and 84.01 for the laptop and the restaurant domain respectively. The B-LSTM \& CRF classifier achieves an \textit{F-score} of 77.96 $\pm$ 0.38 for laptops and an \textit{F-score} of 84.12 $\pm$ 0.2 for restaurants with a confidence interval of 95\%. With our performance, we would have surely won in the laptop domain and probably also in the restaurant domain.

\subsection{Experiments for Unsupervised ATE}
We also perform experiments for ATE with unsupervised learning. For training, we use the automatically labelled datasets (hereafter denoted as ALD) obtained using the methodology described in Section~\ref{ssec:data labelling} with $q_{th}=0.7$ and $q_{th}=0.6$ for the laptop and the restaurant domain respectively. For testing, we use the human labelled datasets (hereafter denoted as HLD) of the SemEval-2014 ABSA task.

\noindent
Our main goal is to evaluate our unsupervised technique on human annotated datasets. To the best of our knowledge, the largest available human annotated datasets for ATE are provided by the SemEval ABSA task and contain laptop and restaurant reviews. Therefore, our analysis focuses only on these two domains.

\noindent
We start by creating a rule-based baseline model to make predictions for the HLD simply by applying techniques presented in Section~\ref{ssec:data labelling}. This baseline (presented in the following section) does not rely on any machine learning techniques for the annotation procedure.

\noindent
We aim at beating the rule-based baseline by using machine learning. To this end, we use the ALD to train our classifier. For unsupervised ATE, we run two types of experiments. The first one uses the traditional IOB labelling format and is stricter. The second one is more relaxed and uses only B and O labels (i.e. I labels are converted to B). The intuition is that aspect terms can be identified by separating B and I labels from O. Therefore, I and B labels are treated equally against O. 

\subsubsection*{Rule-based Baseline Model} \label{sssec: baseline model}
The methodology described in Section~\ref{ssec:data labelling} is used in order to make predictions on the HLD for laptops and restaurants, i.e. the rule-based baseline model does not use any machine learning algorithm. During the annotation process, a token of the HLD is labelled as a target if 
\begin{inlinelist}
    \item it belongs in the pruned quality phrases list and
    \item satisfies at least one of the rules in Table~\ref{tab:syntactic rules}.
\end{inlinelist}
A comparison between the predicted and the golden labels of the HLD gives a quality estimation of the syntactic rules we create and acts as a baseline.

\subsubsection*{SVM} \label{sssec: svm}
We train a linear SVM classifier\footnote{We use the implementation of \texttt{LIBLINEAR}~\cite{liblinear}.} in order to create a second baseline model that uses machine learning. For SVM, we use the baseline features presented in~\cite{stratos} and build 1-0 feature vectors by exploiting the word morphology and the sentence structure (i.e. adjacent words of each token). The training and the evaluation are done using the ALD and the HLD respectively.

\noindent
In addition, we wish to show the trade-off between precision and recall for different values of $q_{th}$. We perform experiments for different values of $q_{th}$ and validate that the higher $q_{th}$ the higher the precision and the lower the recall. For example, an SVM classifier trained on an ALD with $q_{th}=0.7$ achieves an $F_1=39.63$ and $P=71.54$ (Table~\ref{tab:unsupervised ate experiments} shows results for $q_{th}=0.6$ for restaurants). We choose to keep $q_{th}=0.6$ for the restaurant domain because we are interested in a good combination of high precision and \textit{F-score}.

\begin{table}[t]
\centering
\resizebox{0.48\textwidth}{!}{%
\begin{tabular}{c|cc|cc|c}
 & \multicolumn{2}{c|}{\textbf{Labels: IOB}} & \multicolumn{2}{c|}{\textbf{Labels: OB}} &  \\ \cline{2-5}
 & \textbf{P} & \textbf{$\mathbf{F_1}$} & \textbf{P} & \textbf{$\mathbf{F_1}$} &  \\ \hline \hline
\begin{tabular}[c]{@{}c@{}}Rule-based\end{tabular} & 65.13 & 24.35 & 76.65 & 23.76 & \multirow{5}{*}{\rotatebox[origin=c]{90}{Laptops}} \\ \cline{1-5}
SVM & 61.64 & 37.94 & 72.02 & 43.29 &  \\ \cline{1-5}
\begin{tabular}[c]{@{}c@{}}B-LSTM\\ \& CRF\end{tabular} & \textbf{66.67} & \textbf{42.09} & \textbf{74.51} & \textbf{44.37} &  \\ \cline{1-5}
\begin{tabular}[c]{@{}c@{}}SemEval\end{tabular} &  & 35.64 &  &  &  \\ \hline \hline
\begin{tabular}[c]{@{}c@{}}Rule-based\end{tabular} & 84.26 & 28.74 & 96.67 & 27.37 & \multirow{5}{*}{\rotatebox[origin=c]{90}{Restaurants}} \\ \cline{1-5}
SVM & 67.28 & 48.08 & 80.83 & 57.36 &  \\ \cline{1-5}
\begin{tabular}[c]{@{}c@{}}B-LSTM\\ \& CRF\end{tabular} & \textbf{74.03} & \textbf{53.93} & \textbf{83.19} & \textbf{63.09} &  \\ \cline{1-5}
\begin{tabular}[c]{@{}c@{}}SemEval\end{tabular} &  & 47.15 &  &  & 
\end{tabular}%
}
\caption{Experiments for unsupervised ATE. We compare B-LSTM \& CRF classifier against the rule-based baseline, an SVM classifier and the baseline of the SemEval-2014 ABSA contest.}
\label{tab:unsupervised ate experiments}
\end{table}

\subsubsection*{B-LSTM \& CRF}
We employ the B-LSTM \& CRF classifier using the ALD as training set and the HLD as test set, i.e. the evaluation is performed on the human annotated datasets of SemEval-2014 ABSA task. In addition, we use the ABSA training sets of SemEval-2014 as validation sets.

\begin{figure}[t]
    \centering
    \begin{subfigure}{0.48\textwidth}
        \includegraphics[width=\textwidth]{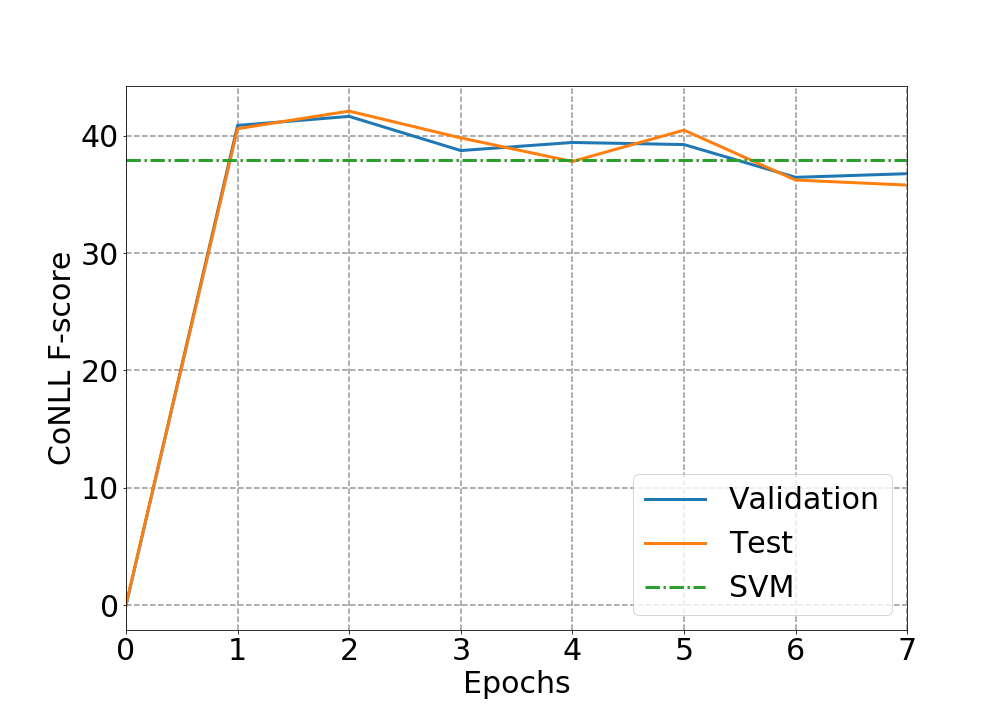}
        \vspace{-\baselineskip}
        %\caption{CoNLL \textit{F-score}}
        \label{laptops_fscore_bio}
    \end{subfigure}
    \vspace{-0.7cm}
    \begin{subfigure}{0.48\textwidth}
        \includegraphics[width=\textwidth]{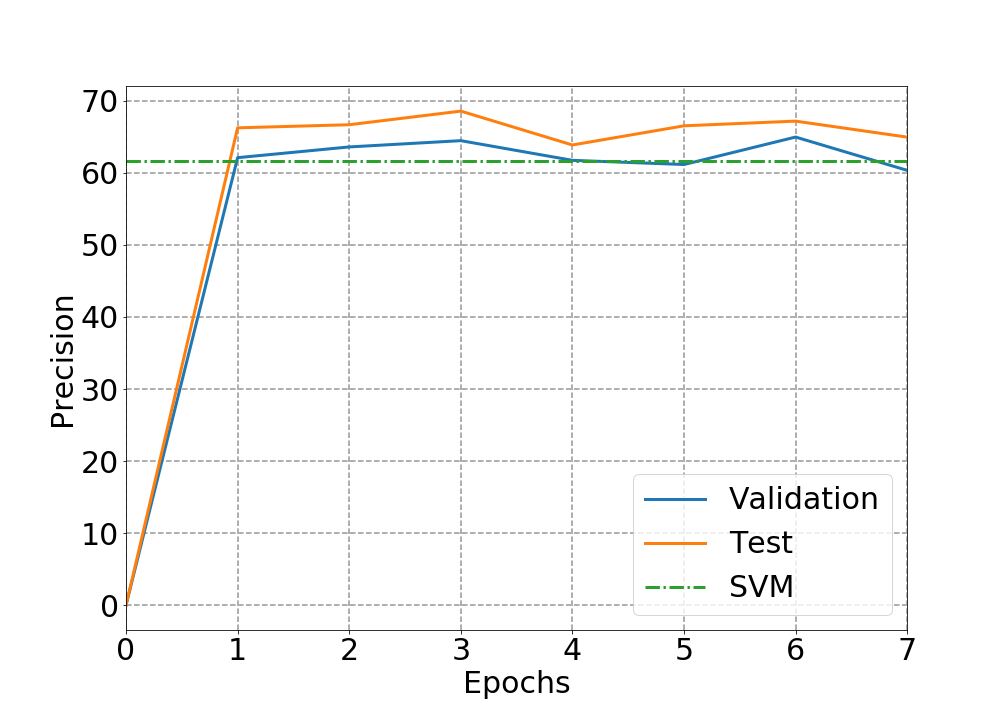}
        %\caption{Precision}
        \label{laptops_precision_bio}
    \end{subfigure}
    \caption{\textit{F-score} (top) and precision (bottom) comparison between B-LSTM \& CRF and SVM for unsupervised ATE in the laptop domain. B, I and O labels are used.} \label{fig:neuroate experiments}
\end{figure}

\noindent
The maximum number of epochs and the patience are set to 20 and 5 respectively. As stopping criterion, we simply choose the epoch that achieves the best \textit{F-score} on the validation set. 
%However, a more sophisticated system could choose the epoch that achieves a good combination\footnote{A function that combines the precision and the \textit{F-score} should be defined (e.g. a mean of the two).} of precision and \textit{F-score}. 
In all our experiments, we compare the performance of the B-LSTM \& CRF classifier with the respective performance of the rule-based baseline and the SVM model. 
\noindent
We do not report any confidence intervals for the B-LSTM \& CRF classifier because the training time increases dramatically in the case of unsupervised ATE due to the increased size of the dataset. Conducting one experiment usually takes more than $15h$, which means that a round of at least 20 experiments, that would allow for defining confidence intervals, would be computationally intensive. For this reason, we leave the report of confidence intervals for unsupervised ATE for future work. However, we repeat up to 3 experiments for each case and verify that the CoNLL \textit{F-score} and the precision are always higher compared to SVM. Results for the laptop domain can be visualized in Fig.~\ref{fig:neuroate experiments}. We do not present any figures for the restaurant domain since the learning curves are very similar to the ones of the laptop domain.

\noindent
We draw several conclusions by observing the results tabulated in Table~\ref{tab:unsupervised ate experiments}. First, the B-LSTM \& CRF classifier achieves higher \textit{F-score} for both domains compared to the rule-based baseline model and the SVM classifier. The \textit{F-score} relative improvement between the rule-based baseline and the B-LSTM \& CRF classifier is 73\% and 88\% for the laptop and the restaurant domain respectively. At the same time, we preserve high precision and attain values above 80\%. Finally, our unsupervised method beats the supervised baseline \textit{F-score} from SemEval ABSA.

%% file: sections/conclusion.tex
We present a B-LSTM \& CRF classifier which we use for feature extraction and aspect term detection for both supervised and unsupervised ATE. We validate this classifier by performing supervised ATE and achieving top-ranking performance on the human annotated datasets of the SemEval-2014 ABSA contest for the laptop and restaurant domain. Moreover, we introduce a new, automated, unsupervised and domain independent method to label tokens of raw opinion texts as aspect terms with high precision. We use the automatically labelled datasets to train the B-LSTM \& CRF classifier, which we evaluate on human annotated datasets. Against all odds, our unsupervised method beats the supervised ABSA baseline \textit{F-score} from SemEval, while preserving high precision scores.

\noindent
As future work, we plan to perform ATE for different domains (e.g. hotels) using our methods. Moreover, we plan to work towards adapting our techniques to multilingual datasets (e.g. French, Spanish, etc.). We would also investigate the idea of exploiting the available ratings (e.g. 1-5 stars) of the review datasets in order to construct new datasets for ATE. This would allow us to perform ATE with distant supervision.